\pdfoutput=1
\documentclass[11pt]{article}

\usepackage[final]{acl}

\usepackage{times}
\usepackage{latexsym}

\usepackage[T1]{fontenc}

\usepackage[utf8]{inputenc}

\usepackage{microtype}

\usepackage{inconsolata}

\usepackage{graphicx}
\usepackage{xcolor}

\usepackage{booktabs}
\usepackage{hyperref}
\usepackage{multirow}
\usepackage{multicol}
\usepackage[most]{tcolorbox}
\usepackage{adjustbox}
\usepackage{graphicx}
\usepackage{fullpage}
\usepackage{times}
\usepackage{fancyhdr,graphicx,amsmath,amssymb}
\usepackage{algorithm}
\usepackage{algpseudocode}
\usepackage{booktabs}
\usepackage{adjustbox}
\usepackage{url}
\usepackage{hyperref}
\usepackage{amssymb}
\usepackage{marvosym}
\usepackage{multirow}
\usepackage{subcaption}

\usepackage{nopageno}

\newtcolorbox{promptbox}[2][]{
  colback=gray!10,
  colframe=gray!50,
  arc=3mm,
  boxrule=1pt,
  left=10pt,
  right=10pt,
  top=8pt,
  bottom=8pt,
  before skip=12pt,
  after skip=12pt,
  fonttitle=\bfseries,
  title=#2,
  #1
}

\title{Quality-Aware Decoding: Unifying Quality Estimation and Decoding}

\author{Sai Koneru$^{1}$,
  Matthias Huck$^{2}$,
  Miriam Exel$^{2}$, \textnormal{and}
  Jan Niehues$^{1}$ \\
  $^{1}$ Karlsruhe Institute of Technology \\
  $^{2}$ SAP SE, Dietmar-Hopp-Allee 16, 69190 Walldorf, Germany \\
  \texttt{\{sai.koneru, jan.niehues\}@kit.edu} \\
  \texttt{\{matthias.huck, miriam.exel\}@sap.com}}

\begin{document}
\maketitle
\begin{abstract}
Quality Estimation (QE) models for Neural Machine Translation (NMT) predict the quality of the hypothesis without having access to the reference.
An emerging research direction in NMT involves the use of QE models, which have demonstrated high correlations with human judgment and can enhance translations through Quality-Aware Decoding. Although several approaches have been proposed based on sampling multiple candidate translations and picking the best candidate, none have integrated these models directly into the decoding process. In this paper, we address this by proposing a novel token-level QE model capable of reliably scoring partial translations. We build a uni-directional QE model for this, as decoder models are inherently trained and efficient on partial sequences. We then present a decoding strategy that integrates the QE model for Quality-Aware decoding and demonstrate that the translation quality improves when compared to the N-best list re-ranking with state-of-the-art QE models (up to $1.39$ XCOMET-XXL $\uparrow$). Finally, we show that our approach provides significant benefits in document translation tasks, where the quality of N-best lists is typically suboptimal\footnote{Code can be found at \url{https://github.com/SAP-samples/quality-aware-decoding-translation}}
\end{abstract}
\section{Introduction}

Large language models (LLMs) have significantly impacted various Natural Language Processing (NLP) tasks \citep{brown2020language, jiang2023mistral, dubey2024llama}, including Neural Machine Translation (NMT). The field of NMT is transitioning from using dedicated encoder-decoder transformers \citep{vaswani2017attention, nllb2024scaling} to leveraging decoder-only LLM-based translation models \citep{kocmi2024findings}. This shift is driven by LLMs' ability to retain knowledge, handle large contexts, and follow instructions, learned during extensive pre-training \citep{xu2024contrastive, alves2024tower}. As a result, LLM-based MT models have achieved state-of-the-art translation quality \citep{kocmi2024findings}.

In parallel, Quality Estimation (QE) has become a well-researched subfield within NMT. QE models are trained to predict the quality of a translation without requiring access to the reference \citep{rei2021references,rei2022cometkiwi}. Interestingly, QE models can achieve performance in assessing translation quality that is comparable to MT evaluation models, which do have access to the reference \citep{zerva2024findings}.

This led to the question: "\textit{Can we integrate QE into the current translation process to improve quality?}" Incorporating QE into NMT offers several benefits. First, having a expert QE model guiding the decoding can further improve the quality. Second, by adapting the QE model with feedback from human annotators, we can generate future translations guided with the newly obtained feedback.

\begin{figure*}[!ht]
\includegraphics[width=\textwidth]{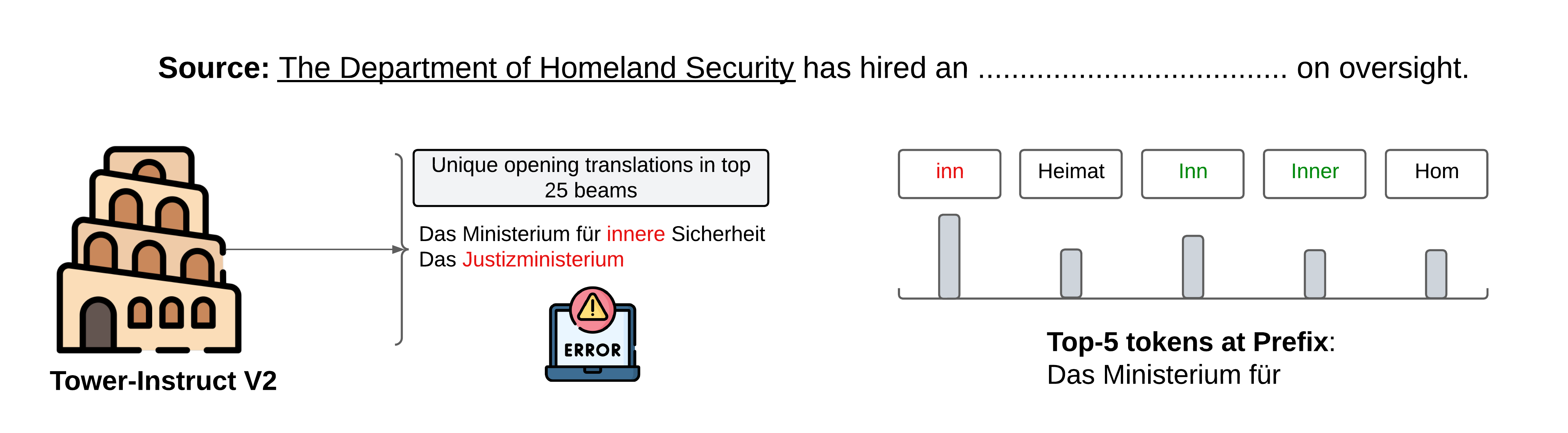}
 \caption{Example from WMT'23 English → German \#ID: 10: The paragraph begins with 'Department of Homeland Security,' which should be translated as 'Ministerium für \textbf{I}nnere Sicherheit.' However, the top 25 beams do not contain the correct translation and begin with an error, making N-best list re-ranking insufficient. Although the top-5 tokens at the decoding contain the correct forms 'Inn' or 'Inner,' the probabilities split among them giving highest mass to the incorrect token 'inn.' Quality-Aware decoding can prevent errors with earlier integration.}
\label{fig:nbestlist}
\end{figure*}

Several approaches have been explored to integrate QE into the translation process. These include re-ranking the N-best list \citep{fernandes2022quality}, applying minimum Bayes risk (MBR) decoding on a quality-filtered N-best list \citep{tomani2024quality}, and training additional models for post-editing based on QE-predicted errors \citep{treviso2024xtower}. However, all these methods operate on fully generated sequences before the QE model can exert influence. Integrating QE earlier in the decoding process, referred in this paper as \textit{Quality-Aware Decoding}, could enhance translation quality and reduce reliance on the N-best list. This is especially relevant when dealing with long inputs as GOOD translations during decoding are likely to be pruned and may need sampling larger number of finished hypothesis. We illustrate this in Figure \ref{fig:nbestlist}.

To achieve this, a QE model capable of predicting the quality of partial translations is required. However, current leading QE models face challenges in this area, as they are typically not trained to predict scores for incomplete hypotheses. \textit{Therefore, developing QE models that can handle partial translations is essential for implementing Quality-Aware Decoding during the translation process}.

In this work, we propose adapting LLM-based NMT models to perform QE on partial translations and incorporating this model into the decoding. We create a token-level synthetic QE dataset using WMT Multidimensional Quality Metrics (MQM) data \citep{burchardt2013multidimensional, freitag2024llms}. We then adapt a uni-directional LLM-based MT model to predict whether a token is \textit{GOOD} or \textit{BAD}. Training QE models on these token-level tasks alleviates the data challenge and allows us to exploit the MQM data while simultaneously making the task easier for the model compared to predicting a score directly.

\begin{figure*}[!ht]
\includegraphics[width=\textwidth]{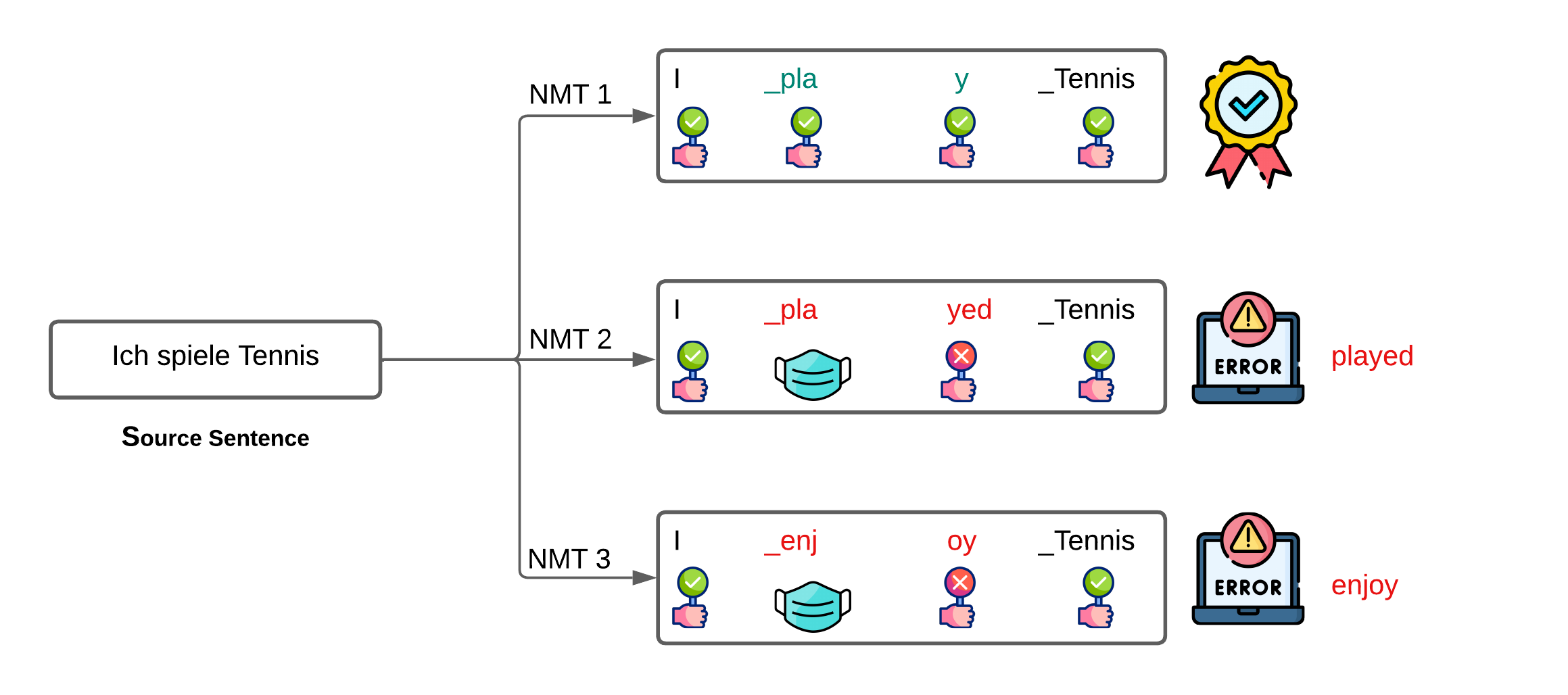}
 \caption{Token-level label annotation scheme using the MQM error tags. \textit{MASK} indicates that this token label will not be used in training to prevent incorrect learning signal.}
\label{fig:annotation}
\end{figure*}

Furthermore, integrating the QE model into NMT during decoding is not trivial, as we need to combine the QE estimates during decoding. Therefore, we modify the decoding strategy from \citet{koneru2024plug} to incorporate token-level predictions efficiently with the adapted QE model to provide real-time feedback during the decoding process. We summarize our contributions below.

\begin{itemize}
    \item We present a novel uni-directional QE model which estimates quality on incomplete hypotheses by averaging the probabilities of each token being classified as \textit{GOOD}. 
    

    \item We propose a decoding strategy that combines the token-level QE model on partial hypothesis and the NMT model to perform Quality-Aware Decoding. 
    
    \item We show through experiments that early integration is essential and the translation quality is improved even when compared to re-ranking the N-best list with state-of-the-art QE models.

    \item We highlight the significance of our approach in document translation scenarios, where post-generation QE techniques fall short due to their reliance on the quality of the N-best list, a challenge that becomes more difficult as the input length increases.
\end{itemize}

\section{Quality-Aware Decoding}

The primary objective of this paper is to achieve Quality-Aware Decoding in NMT. To accomplish this, it is essential to predict the quality of partial translations and integrate this information during the decoding process. Our approach proposes using one NMT model for generating translations and another adapted NMT model to predict the quality of the candidate translations produced by the first model.

First, we explain why relying solely on the NMT model to predict the quality of a hypothesis is insufficient and why an additional model is necessary. Next, we outline the adaptation of the NMT model for QE on partial translations, detailing the creation of a token-level QE dataset, the modifications made to the NMT model for this task, and the process of estimating the sentence-level quality score. Finally, we describe the algorithm used to incorporate the QE score into the decoding process.

\subsection{Decomposing Decoding: Translation + QE}
NMT models generate a token-by-token sequence and provide the probability of each token at the decoding step. The average of the log-probabilities is often used as a proxy to score the candidate during search. 

While NMT models are capable of generating high-quality translations, using the average log-probabilities of hypotheses as a scoring metric tends to yield poor correlation with actual translation quality \citep{eikema2020map, freitag2020bleu}. In many cases, a translation can continue in several different ways, all of which may be acceptable. If the starting tokens for these continuations differ, the probability mass may be spread across multiple options which is used during the search. However, from a quality perspective, all these continuations could still achieve a high score, as the QE scores are independent and need not sum to $1$.

Therefore, we propose a expert model that focuses on quality to estimate the scores better during decoding and  improve the search space leading to a better hypothesis.


\subsection{Quality Estimation on Partial Sequences}

To provide a quality score during decoding, the QE model must be capable of handling incomplete sequences. It should not penalize a sequence if there is a potential extension that could lead to a perfect translation.

Estimating the score in this way is not feasible with current QE models, such as COMET \citep{rei2021references}, as they were not trained for this specific task and cannot provide reliable scores in the context of partial translations. Hence, we need to develop a partial QE system.

When building a partial QE system, several factors need to be considered. First, should the model use a uni-directional or bi-directional architecture? A \textbf{uni-directional} model is more efficient, as it allows for caching the hidden states, which can then be used for subsequent steps without re-encoding, unlike a bi-directional model.

Next, we need to decide whether to predict the QE score at the sequence level or at the token level. For \textbf{token-level QE}, we can directly use data from MQM annotations, as we already know which tokens are \textit{GOOD} or \textit{BAD}. However, for segment-level scoring, we need to consider how to synthetically create the training data. 




Therefore, we decide adapt the uni-directional model into a token-level QE system that predicts whether each token is \textit{GOOD} or \textit{BAD} (a binary decision) by adding an additional classifier head. This adaptation enables us to estimate the score for a sequence by calculating the average probability that each token is classified as \textit{GOOD}. We hypothesize that adapting the model in this way, rather than directly predicting the score, provides greater stability, as the last hidden states inherently contain token-level information and do not require mapping the entire sequence to a single score.

For training this model, we leverage the WMT MQM data containing error annotations in NMT outputs. We can treat tokens before an error as \textit{GOOD} and those containing inside an error as \textit{BAD}. Then, we can train in uni-directional manner where each token's label is predicted using only the preceding context in the hypothesis. This is crucial as we only have the preceding context to estimate the quality for partial hypothesis.

\subsubsection{Learning the Right Signal}

\begin{algorithm*}[!t]
\caption{Computing merged score of partial hypothesis with translation and token-level QE models.}
\begin{algorithmic}[1]
\setlength{\baselineskip}{1.2em}
\Procedure{MergeScore}{}
    \State \textbf{Input:}   Hypothesis tokens $h_1, h_2, h_3, \dots, h_{n}$, Translation Model $\mathcal{M}_{NMT}$, QE model $\mathcal{M}_{QE}$, Source sentence $\mathcal{S}$, Re-ranking weight $\alpha$,
    \State \textbf{Output:} $merged\_score$
    \State $Score_{NMT} \gets \frac{1}{n}\sum \log \mathcal{P}(h_1, h_2,\dots, h_{n}|\mathcal{S};\mathcal{M}_{NMT})$ 
    \State $Score_{QE} \gets \frac{1}{n}\sum \log \mathcal{P}(0_{1}, 0_{2},\dots,0_{n} | h_1, h_2,\dots, h_{n},\mathcal{S};\mathcal{M}_{QE})$ 
    \State $merged\_score \gets (\alpha) \times Score_{NMT} + (1 - \alpha) \times Score_{QE}$
\EndProcedure
\end{algorithmic}
\label{alg:joint}
\end{algorithm*}

The straightforward approach to creating labels is to assign $1$ to all tokens within the error span and $0$ otherwise. However, MQM annotations can mark errors from words to phrases, and the starting tokens of an error span may not always be wrong. This is illustrated in Figure \ref{fig:annotation}.

For example, consider the German sentence \textit{"Ich spiele Tennis"} translated by three different NMT systems, each annotated with MQM error labels. In this work, we focus on learning a binary decision: whether an error is present, ignoring error severity.

\textbf{System 1: No error}: The translation \textit{"I play Tennis"} is perfect, and all tokens are labeled as "\textit{GOOD}."

\textbf{System 2: Partial error}: The translation \textit{"I played Tennis"} has an error in the verb form ("played" instead of "play"). The error is in the token span \textit{"played"}, but not all tokens in this span are incorrect (e.g., "pla" is correct). Assigning a "\textit{BAD}" label to the entire span would lead to incorrect learning. A more refined approach is needed to mark errors accurately at the token level.

\textbf{System 3: Full error}: The translation \textit{"I enjoy Tennis"} contains an error in \textit{"enjoy"}, so all tokens in this span should be labeled as "\textit{BAD}."

It is not trivial to decide when the prefix of an error span is correct/incorrect. To achieve accurate labeling, we propose the following scheme:

\begin{itemize} \item Apply a \texttt{<MASK>} operation to all tokens within the error span. \item Only the last token in the span is assigned the label "\textit{BAD}", as the error is considered complete at the end of the span. \end{itemize}

If the error token is in the middle, we still train the model to predict "\textit{BAD}" in the end and let the model determine which tokens should be part of the error span during inference. This approach ensures that errors are identified without explicitly defining the error span. 

\subsubsection{Sequence-Level Quality Estimation}

After fine-tuning a token-level classification model to predict the quality of the tokens, we still need to map these predictions into a sequence-level score that can be integrated during the decoding process. There are several potential ways to achieve this.

One approach is to simply count how many tokens are classified as \textit{BAD} in the current hypothesis. However, this method has limitations. The number of errors should be normalized based on the length of the hypothesis to account for varying sizes. Additionally, converting the probabilities into a fixed number of error tokens would need to account for different error types according to the MQM format, as each error counts differently.

To avoid such strict scoring schemes, we take a simpler approach. We average the log probabilities of all tokens that are classified as \textit{GOOD}. This method inherently accounts for the length of the hypothesis, and it provides a score on the scale of log probabilities, which aligns with the decoding process. Therefore, we use this averaged log probability as a proxy metric for the QE score, where a higher score indicates better quality
(\textbf{Line 5} in Algorithm \ref{alg:joint}).

\subsubsection{Fusing Translation and Quality}

We can use a token-level QE system to evaluate the quality of a source and partial hypothesis during decoding. However, integrating these probabilities into all candidates is computationally expensive, as each beam considers extensions equal to the vocabulary size.

To address this, we adopt a simplified decoding strategy from \citet{koneru2024plug}, which ensembles models with different vocabularies. By adapting the same MT model for token-level QE, we simplify the merging process, as the vocabularies match. This restriction is reasonable, as it is also beneficial to leverage the knowledge learned by the specialized MT for token-level QE.

The core idea is to re-rank the top candidates at each decoding step using the QE model. After re-ranking, the translation and QE scores are merged, and the process repeats until the end-of-sentence token is generated, for each beam. This strategy allows us to efficiently incorporate the QE model’s estimate, improving translation quality.

During decoding, at each step, we have scores for $n$ beams and $V$ possible extensions from the vocabulary. In typical beam search, we select the top $n$ extensions and expand the hypothesis. To make the decoding process Quality-aware, we estimate the quality of these extensions. Since estimating all extensions is computationally expensive, we limit the candidates by selecting a specified number of top candidates.

To achieve this, we use a hyper-parameter $topk$, which selects the best $topk$ extensions for each beam. For each of these top $topk$ extensions, we compute a combined score, detailed in Algorithm \ref{alg:joint}. This combined score incorporates both the translation model score and the quality estimation score, ensuring the quality is considered during decoding.

For a top extension at decoding step $n$, let the current tokens be $h_1, h_2, h_3, \dots, h_n$. The NMT model score is computed as the average log probabilities of each token (Line 4). For the token-level QE model, we compute the average probability of each token being classified as '\textit{GOOD}' (Line 5). The merged score is equal to weighted linear combination of these probabilities, with weight $\alpha$ (Line 6).

Thus, to make the decoding process Quality-Aware, we first train a token-level QE system by adapting the same NMT model to ensure vocabulary matching. We then combine the scores from both models to improve the sequence estimates explored during search.

\begin{table}[!ht]
\resizebox{\columnwidth}{!}{
\centering
\begin{tabular}{@{}c|ccc@{}}
\toprule
                                                                                      & Pearson        & Spearmann      & Kendall        \\ \midrule
COMETQE                                                                               & \textbf{44.41} & 41.29          & 31.19          \\ \midrule
COMETQE-XL                                                                            & 41.23          & \textbf{42.17} & \textbf{31.84} \\ \midrule
Tower Avg. Log Prob                                                                        & 32.32          & 16.74          & 12.77          \\ \midrule
\begin{tabular}[c]{@{}c@{}}Tower QE\end{tabular} & 40.56          & 33.96          & 25.87          \\ \bottomrule
\end{tabular}
}
\caption{Correlation on WMT 23 for English $\rightarrow$ German Test set. The scores are calculated after removing the few sentences labeled for hallucination detection. Best scores according to each coefficient are highlighted in \textbf{bold}.}
\label{tab:correlation}
\end{table}

\begin{table*}[!ht]
\resizebox{2\columnwidth}{!}{
\begin{tabular}{@{}ccccc@{}}
\toprule
\multicolumn{1}{c|}{Model}            & \multicolumn{1}{c|}{Beams}                & \multicolumn{1}{c|}{Re-ranking}              & MetricX ($\downarrow$)     & XCOMET-XXL ($\uparrow$)    \\ \midrule
\multicolumn{5}{c}{\textit{English $\rightarrow$ German}}                                                                                                          \\ \midrule
\multicolumn{1}{c|}{Tower}            & \multicolumn{1}{c|}{5}                    & \multicolumn{1}{c|}{\_}                      & 2.52          & 86.93          \\
\multicolumn{1}{c|}{Tower}            & \multicolumn{1}{c|}{25}                   & \multicolumn{1}{c|}{XCOMET-XL QE}            & 2.37          & 87.79          \\
\multicolumn{1}{c|}{Tower}            & \multicolumn{1}{c|}{25}                   & \multicolumn{1}{c|}{Tower QE} & 2.38          & 87.40          \\
\multicolumn{1}{c|}{Tower + Tower QE} & \multicolumn{1}{c|}{5 (25* for Tower QE)} & \multicolumn{1}{c|}{\_}                      & 2.12          & 88.95*          \\
\multicolumn{1}{c|}{Tower + Tower QE} & \multicolumn{1}{c|}{5 (25* for Tower QE)} & \multicolumn{1}{c|}{XCOMET-XL QE}            & \textbf{2.09*} & \textbf{89.08*} \\ \midrule
\multicolumn{5}{c}{\textit{Chinese $\rightarrow$ English}}                                                                                                         \\ \midrule
\multicolumn{1}{c|}{Tower}            & \multicolumn{1}{c|}{5}                    & \multicolumn{1}{c|}{\_}                      & 2.42          & 88.91          \\
\multicolumn{1}{c|}{Tower}            & \multicolumn{1}{c|}{25}                   & \multicolumn{1}{c|}{XCOMET-XL QE}            & 2.30          & 89.49          \\
\multicolumn{1}{c|}{Tower}            & \multicolumn{1}{c|}{25}                   & \multicolumn{1}{c|}{Tower QE} & 2.32          & 89.51          \\
\multicolumn{1}{c|}{Tower + Tower QE} & \multicolumn{1}{c|}{5 (25* for Tower QE)} & \multicolumn{1}{c|}{\_}                      & 2.26          & 89.82*          \\
\multicolumn{1}{c|}{Tower + Tower QE} & \multicolumn{1}{c|}{5 (25* for Tower QE)} & \multicolumn{1}{c|}{XCOMET-XL QE}            & \textbf{2.24*} & \textbf{90.00*} \\ \bottomrule
\end{tabular}
}
\caption{Translation Quality on WMT23 English $\rightarrow$ German Test set. Both XCOMET and MetricX columns use reference for reporting translation quality where as XCOMET-XL QE does not use for re-ranking. Best scores according to each metric are highlighted in \textbf{bold}. We report the top cluster indicated with asterisks and are no worse than other systems determined by Paired T-Test and bootstrap resampling with p < 0.05}
\label{tab:qadecoding}
\end{table*}

\section{Experimental Setup}
\paragraph{Datasets:} We focus on two language directions given their availability of MQM data: English $\rightarrow$ German and Chinese $\rightarrow$ English. To train our token-level QE systems, we use the MQM datasets\footnote{https://github.com/google/wmt-mqm-human-evaluation} from WMT \citep{freitag2021experts}. Specifically, we use the datasets until 2022 for training, 2024 for validation, and 2023 for testing \citep{kocmi2024findings}. This setup is consistent with all the other QE metrics, and we do not use any additional data beyond these datasets.
\vspace{-0.1cm}
\paragraph{Models:} 
Our proposed approach achieves Quality-Aware decoding by combining an NMT model with a token-level QE model, where we adapt the same NMT for QE by adding a classification head. We use the state-of-the-art NMT model, Tower 7B\footnote{Unbabel/TowerInstruct-7B-v0.2} \citep{alves2024tower}, which provides high-quality translations and has already been exposed to MQM data during instruction-tuning. This ensures that the gains observed in our approach stem from integrating Quality-Aware decoding into the NMT process, rather than introducing new data. We find $\alpha$ by setting it to the optimal re-ranking weight on the validation set ( See Appendix \ref{sec:robustalpha} for details). Additional details on training the QE models and hyper-parameters during decoding are provided in Appendix \ref{sec:training_detail}.

\paragraph{Metrics:}
For reporting the translation quality, we consistently use XCOMET-XXL\footnote{Unbabel/XCOMET-XXL} \citep{guerreiro2024xcomet} and MetricX\footnote{google/metricx-24-hybrid-xl-v2p6} \citep{juraska2024metricx} \textbf{with the reference}. To compare with N-best list re-ranking, we use the XCOMET-XL QE\footnote{Unbabel/XCOMET-XL} \textbf{without the reference}. This approach allows us to avoid biasing toward a single metric during the re-ranking process and enables us to measure the gains achieved by differently trained metrics. 

\section{Results}

\begin{table*}[!ht]
\centering
\resizebox{2\columnwidth}{!}{
\begin{tabular}{@{}ccccc@{}}
\toprule
\multicolumn{1}{c|}{Model}            & \multicolumn{1}{c|}{Beams}                        & \multicolumn{1}{c|}{Re-ranking}               & MetricX ($\downarrow$) & XCOMET-XXL ($\uparrow$) \\ \midrule
\multicolumn{5}{c}{\textit{English $\rightarrow$ German}}                                                                                                          \\ \midrule
\multicolumn{1}{c|}{Tower}            & \multicolumn{1}{c|}{25}                           & \multicolumn{1}{c|}{XCOMET-XL QE}             & 2.37     & 87.79      \\
\multicolumn{1}{c|}{Tower}            & \multicolumn{1}{c|}{25}                           & \multicolumn{1}{c|}{Tower QE}         & 2.38     & 87.40      \\
\multicolumn{1}{c|}{Tower}            & \multicolumn{1}{c|}{25}                           & \multicolumn{1}{c|}{Tower Distill QE} & 2.38     & 87.39      \\
\multicolumn{1}{c|}{Tower + Tower QE} & \multicolumn{1}{c|}{5 (25* for Tower QE)}         & \multicolumn{1}{c|}{\_}                       & 2.12     & \textbf{88.95}      \\
\multicolumn{1}{c|}{Tower + Tower QE} & \multicolumn{1}{c|}{5 (25* for Tower Distill QE)} & \multicolumn{1}{c|}{\_}                       & \textbf{2.11}     & 88.76      \\ \bottomrule
\end{tabular}
}
\caption{Performance of Unidirectional QE trained with/without distillation on WMT23 English $\rightarrow$ German Test set. Best scores according to each metric are highlighted in \textbf{bold}.}
\label{tab:towerdistill}
\end{table*}

\begin{table*}[!ht]
\centering
\resizebox{2\columnwidth}{!}{
\begin{tabular}{@{}cccccc@{}}
\toprule
\multicolumn{1}{c|}{Model}            & \multicolumn{1}{c|}{Beams}                & \multicolumn{1}{c|}{Re-ranking}       & XCOMET-XL ($\uparrow$)     & \multicolumn{1}{c|}{XCOMET-XXL ($\uparrow$)}     & Impact                                                                                       \\ \midrule
\multicolumn{6}{c}{\textit{Paragraph-Level}}                                                                                                                                                                                                                                    \\ \midrule
\multicolumn{1}{c|}{Tower}            & \multicolumn{1}{c|}{25}                   & \multicolumn{1}{c|}{XCOMET-XL QE}     & \textbf{86.56} & \multicolumn{1}{c|}{87.79}          & \multirow{3}{*}{\begin{tabular}[c]{@{}c@{}}$\delta$ = + 1.16\\ (88.95 - 87.79)\end{tabular}} \\
\multicolumn{1}{c|}{Tower}            & \multicolumn{1}{c|}{25}                   & \multicolumn{1}{c|}{Tower QE} & 85.40          & \multicolumn{1}{c|}{87.40}          &                                                                                              \\
\multicolumn{1}{c|}{Tower + Tower QE} & \multicolumn{1}{c|}{5 (25* for Tower QE)} & \multicolumn{1}{c|}{\_}               & 86.36          & \multicolumn{1}{c|}{\textbf{88.95}} &                                                                                              \\ \midrule
\multicolumn{6}{c}{\textit{Sentence-Level}}                                                                                                                                                                                                                                     \\ \midrule
\multicolumn{1}{c|}{Tower}            & \multicolumn{1}{c|}{25}                   & \multicolumn{1}{c|}{XCOMET-XL QE}     & \textbf{86.42}          & \multicolumn{1}{c|}{87.68}          & \multirow{3}{*}{\begin{tabular}[c]{@{}c@{}}$\delta$ = + 0.38\\ (88.06 - 87.68)\end{tabular}} \\
\multicolumn{1}{c|}{Tower}            & \multicolumn{1}{c|}{25}                   & \multicolumn{1}{c|}{Tower QE} & 85.23          & \multicolumn{1}{c|}{87.41}          &                                                                                              \\
\multicolumn{1}{c|}{Tower + Tower QE} & \multicolumn{1}{c|}{5 (25* for Tower QE)} & \multicolumn{1}{c|}{\_}               & 85.96          & \multicolumn{1}{c|}{\textbf{88.06}}          &                                                                                              \\ \bottomrule
\end{tabular}
}
\caption{Impact of integrating Unidirectional QE during decoding with paragraphs vs sentences on WMT23 English $\rightarrow$ German Test set. $\delta$ denotes the improvement in translation quality from re-ranking N-best list with XCOMET-XL QE to integrating unidirectional Tower QE during the decoding. Best scores according to each metric are highlighted in \textbf{bold}.}
\label{tab:sentvspara}
\end{table*}

We conduct a series of experiments to validate the effectiveness of Quality-Aware decoding and identify the scenarios where it provides the most benefit. First, we evaluate whether our token-level QE model can better estimate sequence quality compared to the log probabilities of the NMT model. Next, we assess the impact of Quality-Aware decoding by comparing it with other approaches to determine if it improves translation quality. We also perform an ablation study to examine whether training the QE model on errors from the same NMT model enhances its performance. Then, we explore the impact of source sentence length to highlight the limitations of N-best list re-ranking. Finally, we compare our proposed approach with existing Quality-Aware decoding strategies and also their inference time to highlight the latency/quality trade-off. 

\subsection{Quality Estimation Performance}

First, we evaluate the agreement between the Tower-based token-level QE model (\textbf{Tower QE}) and human scores for a given hypothesis. It is only beneficial if we achieve higher correlation than the average of the NMT model log probabilities to show the need to integrate it during decoding. Therefore, we report the correlation with human scores of different models on WMT 23 English $\rightarrow$ German in Table \ref{tab:correlation}. 

We observe that the best-performing systems are the Comet QE models, which predict a single score using the full hypothesis. This is expected, as these models assess quality after the hypothesis is fully generated. In contrast, both log probabilities and Tower QE scores are based on the predicted token of each decoding step, using only the preceding context. Log probabilities perform poorly in this setup, while our proposed model, Tower QE, achieves twice the correlation with human judgments compared to log probabilities, despite scoring token by token with preceding context. This result highlights the potential of integrating our approach into the decoding process.

\subsection{Unified Decoding for NMT}

To validate the effectiveness of our unified decoding approach, we compare it with several baselines in Table \ref{tab:qadecoding}. First, we evaluate whether our approach outperforms generating translations with the NMT model alone. Next, we check if the quality of translations improves compared to N-best list re-ranking. To make the setups comparable, we set $topk$ and $num\_beams$ to $5$ and compare with re-ranking the top $25$ beams using XCOMET-XL. Finally, to demonstrate that re-ranking the N-best list remains a viable and complementary approach, we re-rank the top $5$ beams obtained from Quality-Aware decoding using the same QE model. 

We find that re-ranking with XCOMET-XL and Tower QE yields similar results, indicating that our partial QE model does not over-fit to any specific metric. Furthermore, we observe that the unified decoding approach outperforms N-best list re-ranking across both metrics in both language pairs. For example, the MetricX score improves from $2.37$ to $2.12$ for English $\rightarrow$ German. Note that Tower has already seen this data during instruction-tuning and the improvement is not from new data but from Quality-Aware decoding. Moreover, re-ranking the top $5$ beams obtained from unified decoding with XCOMET-XL leads to a slight further improvement in quality. This highlights the robustness and generalizability of our approach across different evaluation metrics.

\subsection{Adapting for Tower Errors}

We use the MQM annotations from WMT to train our Tower QE model, which contains error annotations from other systems. However, a viable alternative would be to adapt Tower QE specifically to the errors it typically makes. To maintain a similar data setup, we first generate translations using Tower on these source sentences. Then, we annotate the generated hypotheses with XCOMET-XL using the reference and fine-tune Tower QE on this synthetic dataset, which we refer to as \textbf{Tower Distill QE}. We evaluate the performance of the new distill QE model and report the results in Table \ref{tab:towerdistill}.

We observe that the distilled QE model performs very similarly to the QE model trained on errors from other systems. This indicates that there was no significant benefit in adapting the QE model to the specific errors typically made by Tower. However, further analysis on larger datasets and different domains is needed to fully validate the effectiveness of the distillation approach as the current synthetic data generated is small.

\subsection{Sentence vs Document-level Translation}

From Table \ref{tab:qadecoding}, we observe that the gains for English $\rightarrow$ German (paragraph-level) are much higher than for Chinese $\rightarrow$ English (sentence-level). We hypothesize that this discrepancy arises from the length of the sentences, as the N-best list re-ranking is likely sufficient for shorter sentences. To confirm this, we take the English paragraphs and split them into sentences using a tokenizer while tracking the paragraph IDs. We then perform the entire decoding process similarly, and later join the sentences back using the paragraph IDs before evaluation. We report the results in Table \ref{tab:sentvspara}.

We define the impact as the improvement in translation quality from re-ranking the N-best list with XCOMET-XL QE to integrating Tower QE. Comparing the results at the paragraph level to those at the sentence level, we observe that the impact decreases, which confirms our hypothesis. Additionally, we obtain better scores at the document level, further highlighting the potential benefits of Quality-Aware Decoding.

\begin{table*}[!ht]
\resizebox{2\columnwidth}{!}{
\begin{tabular}{@{}c|cccc@{}}
\toprule
\textit{Model}                      & \textit{Beams} & \begin{tabular}[c]{@{}c@{}}Re-ranking/\\ Utility Metric\end{tabular} & \textit{MetricX} & \textit{XCOMET-XXL} \\ \midrule
Tower                               & 25             & XCOMET-XL                                                            & 2.37             & 87.79               \\ \midrule
Tower + MBR                           & 25             & wmt22-comet-da                                                       & 2.75             & 87.53               \\ \midrule
Tower + QE-Fusion                     & 25             & XCOMET-XL                                                            & 2.38             & 87.76               \\ \midrule
Tower + Tower QE (Ours)             & 5(25)          & \_                                                                   & 2.12             & 88.95*              \\ \midrule
Tower + Tower QE (Ours)             & 5(25)          & XCOMET-XL                                                            & \textbf{2.09*}   & \textbf{89.08*}     \\ \midrule
Tower + Tower QE (Ours) + QE-Fusion & 5(25)          & XCOMET-XL                                                            & 2.10             & 89.03*              \\ \bottomrule
\end{tabular}
}
\caption{Translation quality on WMT 23 English $\rightarrow$ German. We report the top cluster indicated with asterisks and are no worse than other systems determined by Paired T-Test and bootstrap resampling with \textbf{p < 0.05}.  For MBR, we use wmt22-comet-da as the utility metric wheras XCOMET-XL for QE-Fusion \citep{vernikos2024don}.}
\label{tab:qasampling}
\end{table*}

\begin{table}[h]
\resizebox{\columnwidth}{!}{
\centering
\begin{tabular}{@{}cccc@{}}
\toprule
\textit{Model}                                                      & \textit{Beams} & \textit{\begin{tabular}[c]{@{}c@{}}Avg Time\\ (Seconds)\end{tabular}} & \textit{XCOMET-XXL} \\ \midrule
Tower                                                               & 5              & 5.20                                                                  & 86.93               \\ \midrule
\begin{tabular}[c]{@{}c@{}}Tower \\  XCOMET-XL Rerank\end{tabular}  & 25             & 13.16 + 1.78                                                          & 87.79               \\ \midrule
\begin{tabular}[c]{@{}c@{}}Tower \\ MBR wmt22-comet-da\end{tabular} & 25             & 13.16 + 0.34                                                          & 87.56               \\ \midrule
\begin{tabular}[c]{@{}c@{}}Tower\\ Tower QE (Ours)\end{tabular}     & 5(25)          & 21.04                                                                 & 88.94               \\ \bottomrule
\end{tabular}
}
\caption{Average inference time on WMT 23 English-German document-level test set. For MBR and re-ranking, $13.16$ is the time used for generating 25 candidates.}
\label{tab:inference_time}
\end{table}

\subsection{Compatibility with Sampling-based Strategies}
Our proposed approach integrates the feedback during decoding time to generate high quality N-best list. Therefore, it can be further combined with strategies such as Minimum Bayes Risk (MBR) \citep{freitag-etal-2022-high} or QE-Fusion \citep{vernikos2024don} decoding that only rely on the sampled candidates. We compare the different decoding strategies and report the scores in Table \ref{tab:qasampling}. For MBR decoding, we do epsilon sampling (epsilon=0.02) and generate 25 candidates with wmt22-comet-da as utility metric. We do not sample more as it is expensive especially at document-level with 7B model and requires multiple runs. For QE fusion, we use XCOMET-XL as the utility metric. We perform this on English-German at paragraph-level as our hypothesis is that the N-best list is problematic for long sequences.

We find that decoding with Tower QE is significantly better than the other approaches in this setting according to XCOMET-XXL metric. We also would like to highlight that it is compatible with sampling based approaches and show that by performing QE-fusion on the top 5 beams with our decoding approach. While the scores are slightly lower, human evaluation is necessary to compare these systems.

\subsection{Inference Time}
While the quality of translation improves, our approach also is computationally more expensive. To demonstrate this, we compare different decoding strategies and report the latency and quality in Table \ref{tab:inference_time}. To calculate the latency, we take the average time in seconds for inference on the WMT 23 English $\rightarrow$ German Test set.

Note that MBR with XCOMET-XL is extremely expensive given the large size of the QE model and the amount of long samples (25*24*557 examples at paragraph level limit batching) that need to be passed through the model due to the exponential nature of MBR. We see that due to the uni-directional nature of the Tower QE, the total inference time is less than double. Further, re-ranking with XCOMET-XL is the fastest given that a single forward pass is needed after generation.

\section{Related Work}

\textbf{Integrating QE in NMT:} Several advancements have been made in improving QE for NMT over the years \citep{rei2021references, rei2022cometkiwi, blain2023findings, zerva2024findings, guerreiro2024xcomet}. These developments have led to the integration of QE in various ways.
One common approach involves applying QE after generating multiple sequences through techniques such as QE re-ranking \citep{fernandes2022quality, faria2024quest} or Minimum Bayes Risk (MBR) decoding \citep{tomani2024quality}. Another direction focuses on removing noisy data using QE models, followed by fine-tuning on high-quality data \citep{xu2024contrastive, finkelstein2024introducing}. \citet{vernikos2024don} proposes to generate diverse translations as a first step and then combine them. We perform this explicitly by integrating the QE directly into decoding.
Recently, \citet{zhang2024learning} exploited the MQM data by training models to penalize tokens within an error span, improving quality. In contrast, our approach adopts a modular framework, where we propose an expert QE model that is trained independently for targeted training. This modular approach aims to improve performance by decomposing the task into separate translation and QE components.

\textbf{Reward Modeling in NLG:}  Quality-Aware decoding shares similarities with controllable text generation, particularly in using a "Quality/Reward" model to guide decoding. Methods like Weighted Decoding \citep{yang2021fudge} adjust token probabilities for controlled generation, while \citet{deng-raffel-2023-reward} use a uni-directional reward model to maintain efficiency. \citet{li-etal-2024-reinforcement} further enhance control with a token-level reinforcement learning-based model. While related, our key contribution is the development of the first uni-directional QE model specifically for translation.

\section{Conclusion}
We demonstrated the value of Quality-Aware decoding in improving translation quality without relying on post-generation methods. Using MQM data, we built a uni-directional token-level QE model and integrated it into the decoding process. Our experiments show measurable quality gains, achieved without adding new training data to the NMT model, highlighting that improvements come only from the decoding approach.

\section{Limitations}
Although our Quality-Aware decoding improves translation quality, it adds considerable computational complexity to the inference process. Theoretically, this approach would double the time required to generate a translation and would require additional memory to utilize the token-level QE model. One potential solution to mitigate this issue could be to use token-level QE as a reward model for training through reinforcement learning.

Furthermore, we trained our model on a limited set of human-annotated MQM data. However, current QE models, such as XCOMET, are capable of predicting error tags using the reference with reasonable quality. This suggests that further improvements could be achieved if these models were trained on larger-scale datasets, providing more nuanced feedback and refining translation quality even further.

In addition, human evaluation is necessary to validate if the translation quality also improves with human judgment. Although we were able to better integrate MQM data during decoding, it decreases confidence in relying completely on automatic metrics.

Lastly, our proposed token-level QE model does not account for error severity. Ideally, it should be able to predict the category of errors, allowing for more nuanced feedback and enabling the model to generate translations with only minor errors when necessary.

\section*{Acknowledgments}
 Part of this work was performed on the HoreKa supercomputer funded by the Ministry of Science, Research and the Arts Baden-Württemberg and by the Federal Ministry of Education and Research.

\bibliography{custom}

\appendix

\section{Appendix}
\label{sec:appendix}

\subsection{Training details}
\label{sec:training_detail}

We use the transformers library \citep{wolf-etal-2020-transformers} for training and inference with Tower-Instruct V2.  For adapting Tower to token-level QE, we use LoRA \citep{hulora} based fine-tuning with an additional classifier head. Therefore, we only train the adapters and the weights for classification head.

We add the adapters to the modules \textit{q\_proj,k\_proj,v\_proj,gate\_proj,up\_proj} and \textit{down\_proj}. We set a batch size for each device to 12 initially and enable \textit{auto\_find\_batch\_size} to \textit{True} on 4 NVIDIA RTX A6000 GPU's. For having a  larger batch size during training, we set \textit{gradient\_accumulation\_steps} to 6. We use a \textit{learning\_rate} of $1e^{-5}$. We set the \textit{eval\_steps} to $50$ and \textit{num\_train\_epochs} to $10$. The other parameters are set to default.

Using the cross-entropy loss for token-level QE directly is insufficient due to the fact that the majority of tokens are classified as '\textit{GOOD}'. Hence, we find that the weighted cross-entropy loss is essential when fine-tuning the model. For the training on human MQM data, we set the weights to $0.05,0.95$ to '\textit{GOOD}' and '\textit{BAD}' labels respectively. In the case of distilling from XCOMET, we observed more errors. Therefore, we find that setting them $0.2,0.8$ to '\textit{GOOD}' and '\textit{BAD}' labels respectively provided stable training.

We train on data until WMT'22 for training and use WMT'24 for validation. We calculate the macro '\textit{F1}' on token-level predictions as the validation metric and stop training if it does not improve for 10 consecutive \textit{eval\_steps}.

\subsection{Partial vs Full Sequence Quality Estimation}

We also compare the difference in performance between our proposed token-level QE for partial sequences with Tower trained for full sequence QE. We achieve this by adding a regression head to predict the score at the end-of-sentence token. Hence, the model uses the source and hypothesis to predict the score using regression head at the end.

We fine-tune the model using only direct assesment data \citep{zerva2024findings} (\textbf{Tower Full DA}). Furthermore, we use this as initialisation and continue fine-tuning on the MQM data (\textbf{Tower Full DA + MQM}). We also use LoRA similarly to the previous model with a regression head to adapt the model. We report the scores in Table \ref{tab:correlation_ablation}.

We see that the both Tower QE models based on full sentences outperforms the partial model. However, this is expected as it has seen the entire context and trained on larger amounts of data. Still, the partial model achieves much higher correlaiton that the log probabilities showcasing its potential for Quality-Aware decoding.

\subsection{Robustness to re-ranking weight}
\label{sec:robustalpha}

We introduce a hyperparameter, $\alpha$, to merge probabilities from the token-level QE model and the translation model. To analyze its impact, we re-rank the N-best list using various $\alpha$ values, avoiding repeated joint decoding. If the QE model is helpful, we expect translation quality to improve when $\alpha < 1$.

\begin{figure*}[!htpb]
    \centering
    \begin{subfigure}[b]{0.5\textwidth}
        \centering
        \includegraphics[width=\textwidth]{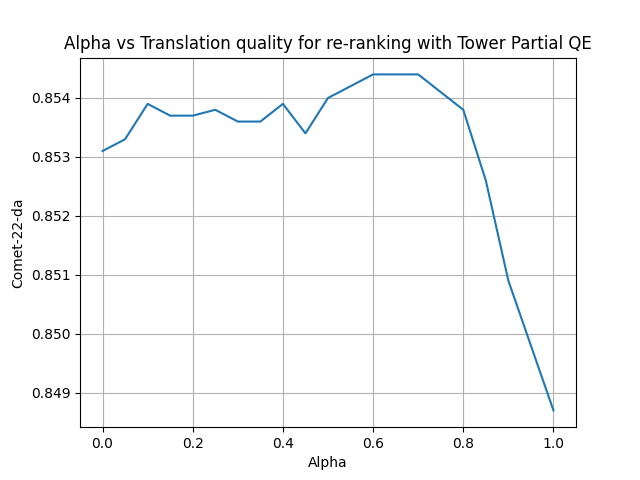} 
        \caption{English $\rightarrow$ German}
        \label{fig:subfigure1}
    \end{subfigure}
    
    \vspace{0.5cm} 

    \begin{subfigure}[b]{0.5\textwidth}
        \centering
        \includegraphics[width=\textwidth]{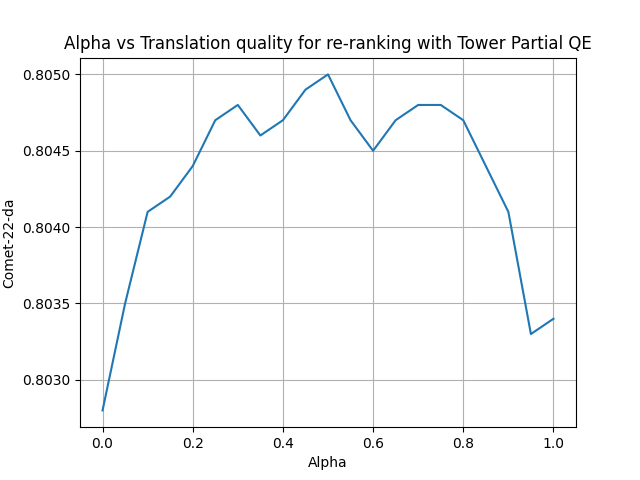} 
        \caption{Chinese $\rightarrow$ English}
        \label{fig:subfigure2}
    \end{subfigure}
    
    \caption{Impact of $\alpha$ when re-ranking with token-level Tower QE on WMT'23 Test sets.}
    \label{fig:mainfigure}
\end{figure*}

Figure \ref{fig:mainfigure} shows that lower $\alpha$ values consistently yield better results, confirming that incorporating QE probabilities improves translation. This highlights the value of Tower QE and shows that re-ranking is an effective and robust way to tune $\alpha$.

\begin{table*}[!htpb]
\centering
\begin{tabular}{@{}c|ccc@{}}
\toprule
                                                                                      & Pearson        & Spearmann      & Kendall        \\ \midrule
COMETQE                                                                               & \textbf{44.41} & 41.29          & 31.19          \\ \midrule
COMETQE-XL                                                                            & 41.23          & \textbf{42.17} & \textbf{31.84} \\ \midrule
\begin{tabular}[c]{@{}c@{}}COMETQE Scratch\\      Fine-tuned (ours)\end{tabular}      & 36.32          & 33.66          & 25.24          \\ \midrule
Tower Log Prob                                                                        & 32.32          & 16.74          & 12.77          \\ \midrule
\begin{tabular}[c]{@{}c@{}}Tower Partial QE\end{tabular} & 40.56          & 33.96          & 25.87          \\ \midrule
Tower Full DA                                                                        & 33.67          & 36.46          & 27.38          \\ \midrule
Tower Full DA + MQM                                                                 & 32.03          & 40.85          & 30.38          \\ \bottomrule
\end{tabular}
\caption{Full Correlation results on WMT 23 for English $\rightarrow$ German Test set. Partial indicates that the QE model predict scores via token-level where as full indicates predicting the score at the end-of-sentence token. The scores are calculated after removing the few sentences labelled for hallucination detection. Best scores according to each coefficient are highlighted in \textbf{bold}.}
\label{tab:correlation_ablation}
\end{table*}

\begin{figure*}[!htpb]
\begin{promptbox}[title={Tower Translation Prompt}]
    \small
    <|im\_start|>user\\
    Translate the sentence from English into German.\\
    English: \{src\_sent\}\\
    German:\\
    <|im\_end|>\\
    <|im\_start|>assistant
\end{promptbox}

\begin{promptbox}[title={Tower Token-Level QE Prompt}]
    \small
    English:\{src\_sent\}\\
    German: \{tgt\_sent\}
\end{promptbox}
\caption{Prompts used in our experiments for translation and QE model. \{src\_sent\} and \{tgt\_sent\} represent the source and target sentence. We replace the language with Chinese and English when experimenting with that language pair.}
\end{figure*}

\end{document}